%% file: main.tex
\documentclass[conference]{IEEEtran}
\IEEEoverridecommandlockouts
\usepackage{cite}
\usepackage{amsmath,amssymb,amsfonts}
\usepackage{algorithmic}
\usepackage{graphicx}
\usepackage{textcomp}
\usepackage{xcolor}
\def\BibTeX{{\rm B\kern-.05em{\sc i\kern-.025em b}\kern-.08em
		T\kern-.1667em\lower.7ex\hbox{E}\kern-.125emX}}
\usepackage{algorithm}
\usepackage{array}
\usepackage[caption=false,font=normalsize,labelfont=sf,textfont=sf]{subfig}
\usepackage{textcomp}
\usepackage{stfloats}
\usepackage{url}
\usepackage{verbatim}
\usepackage{graphicx}
\usepackage{caption}
\usepackage{shortcuts}
\usepackage{comment}
\usepackage{makecell} 	
\usepackage{mathrsfs} 	
\usepackage{amssymb} 	

\usepackage{amsthm}		
\usepackage{multirow}	
\usepackage{enumitem}	
\usepackage{pifont}		

\usepackage[misc,geometry]{ifsym}
\usepackage{array}		
\newcolumntype{H}{>{\setbox0=\hbox\bgroup}c<{\egroup}@{}}

\theoremstyle{definition}
\newtheorem{definition}{Definition}[section]

\usepackage{dsfont}

\makeatletter
\newcommand{\linebreakand}{%
\end{@IEEEauthorhalign}
\hfill\mbox{}\par
\mbox{}\hfill\begin{@IEEEauthorhalign}
}
\makeatother

\begin{document}
	
	\title{Supervised Robustness-preserving Data-free\\Neural Network Pruning}
	
	\author{\IEEEauthorblockN{Mark Huasong Meng\IEEEauthorrefmark{1}\IEEEauthorrefmark{2},
			Guangdong Bai\IEEEauthorrefmark{3}\IEEEauthorrefmark{1}\textsuperscript{(\Letter)}\thanks{\Letter\, G. Bai and S. G. Teo are corresponding authors.}, Sin G. Teo\IEEEauthorrefmark{2}\textsuperscript{(\Letter)} and
			Jin Song Dong\IEEEauthorrefmark{1}}
		\IEEEauthorblockA{
			\IEEEauthorrefmark{1} National University of Singapore, Singapore,
			\IEEEauthorrefmark{2} Institute for Infocomm Research, A*STAR, Singapore\\
			\IEEEauthorrefmark{3} The University of Queensland, Australia\\
			Email: \{menghs, teo\_sin\_gee\}\@i2r.a-star.edu.sg, g.bai@uq.edu.au, dcsdjs@nus.edu.sg}
	}

	\maketitle

	\begin{abstract}
		\input{sections/abstract}
	\end{abstract}

	\begin{IEEEkeywords}
		Neural networks, pruning, model optimization, robustness.
	\end{IEEEkeywords}

	\section{Introduction}
	\label{sec:intro}
	\input{sections/intro1}

	\section{Background}
	\label{sec:background}
	\input{sections/background}

	\section{Problem Definition}
	\label{sec:problem}
	\input{sections/problem}

	\section{Approach Overview}
	\label{sec:overview}
	\input{sections/overview}

	\section{Supervised Data-free Pruning}
	\label{sec:methodology}
	\input{sections/stochastic_sampling}

	\section{Evaluation}
	\label{sec:experiment}
	\input{sections/experiment1}

	\section{Threats to Validity}
	\input{sections/limitations}

	\label{sec:limitations}
	
	\section{Related Work}
	\label{sec:review}
	\input{sections/review}

	\section{Conclusions}
	\label{sec:conclusion}
	\input{sections/conclusion}

	\bibliographystyle{IEEEtran}
	\bibliography{reference}

\end{document}

%% file: sections/abstract.tex
When deploying pre-trained neural network models in real-world applications, model consumers often encounter resource-constraint platforms such as mobile and smart devices.
They typically use the pruning technique to reduce the size and complexity of the model, generating a lighter one with less resource consumption. Nonetheless, most existing pruning methods are proposed with a premise that the model after being pruned has a chance to be fine-tuned or even retrained based on the original training data.
This may be unrealistic in practice, as the data controllers are often reluctant to provide their model consumers with the original data.

In this work, we study the neural network pruning in the \emph{data-free} context, aiming to yield lightweight models that are not only accurate in prediction but also robust against undesired inputs in open-world deployments.
Considering the absence of fine-tuning and retraining that can fix the mis-pruned units, we replace the traditional aggressive one-shot strategy with a conservative one that treats model pruning as a progressive process.
We propose a pruning method based on stochastic optimization that uses robustness-related metrics to guide the pruning process.
Our method is evaluated with a series of experiments on diverse neural network models.
The experimental results show that it significantly outperforms existing one-shot
data-free pruning approaches in terms of robustness preservation and accuracy.

%% file: sections/intro1.tex
Nowadays deep learning is increasingly applied in solving complex real-world problems, such as cybersecurity~\cite{ling2022they,wang2021intra,feng2022detecting} and sport analytic~\cite{jiang2020deep,dong2015sports}.
Deep learning is usually realized by a \emph{neural network} model that is trained with a large amount of data. Compared with other machine learning models such as linear models or support vector machine (SVM) models, neural networks, or more specifically deep neural networks, are empirically proven to gain an advantage in handling more complicated tasks due to their superior capability to precisely approximate an arbitrary non-linear computation. 

In order to achieve a favorable accuracy and generalization, the common practice to train a neural network is to initialize a model that is large and deep in size.
This causes the contemporary models over-parameterized.
For example, many models in image classification or natural language processing contain millions or even billions of trainable parameters~\cite{krizhevsky2012imagenet,tolstikhin2021mlpmixer}.
Deploying them on resource-constraint platforms, such as the Internet of Things~(IoT) or mobile devices, thus become challenging. 
To resolve this issue,
the \emph{neural network pruning} technique is extensively used. It aims to remove parameters that are  redundant or useless, so as to reduce the model size as well as the demand for computational resources.

Most existing research on model pruning assumes the pruning is performed by the model owner who has the original training dataset.
The majority of existing pruning techniques are discussed with a premise that the models after being pruned are going to be fine-tuned or even retrained using the original dataset~\cite{wang2020pruning,molchanov2017pruning,luo2017thinet,suau2020filter}.
As a result, they tend to use aggressive and coarse-grained \emph{one-shot} pruning strategy with the belief that the mis-pruned neurons, if any, could be fixed by fine-tuning and retraining.

This strategy, however, seriously compromises the applicability of pruning.
In practice, the model pruning is mostly performed by the model consumers to adapt the model for the actual deployment environment.
We refer to this stage as the \emph{deployment stage}, to differentiate it from the training and tuning stages occurring at the data controller side~\cite{guo2019empirical}.
In the deployment stage, the model consumers typically have no access to the original training data that are mostly private and proprietary~\cite{taigman2014deepface}.
In addition, data controllers even have to refrain from providing their data due to strict data protection regulations like the EU General Data Protection Regulation (GDPR)~\cite{gdpr2016}.
Therefore, pruning without its original training data, which we refer to as \emph{data-free pruning}, is desirable.

In this work, we approach this problem through the lens of software engineering methodologies.
To address the challenge of the lack of post-pruning fine-tuning,
we design our pruning as a supervised iterative and progressive process, rather than in a one-shot manner.
In each iteration, it cuts off a small set of units and evaluates the effect, so that the mis-pruning of units that are crucial for the network's decision making can be minimized.
We propose a two-stage approach to identify the units to be cut off.
At the first stage, it performs a candidates prioritizing based on the relative significance of the units. At the second stage, it carries out a stochastic sampling with the simulated annealing algorithm~\cite{van1987simulated}, guided by metrics quantifying the desired property.
This allows our method to prune the units that have a relatively low impact on the property, and eventually approaches the optimum.

Our pruning method is designed to pursue \emph{robustness preservation}, given that the model may be exposed to unexpected or even adversarial inputs~\cite{goodfellow2015explaining,madry2018towards,li2021deeppayload} after being deployed in a real-world application scenario.
Our solution is to encode the robustness as metrics and embed them into the stochastic sampling to guide the pruning process.
It stems from the insight that a small and uniformly distributed pruning impact on each output unit is favored to preserve the robustness of the pruned model.
We use two metrics to quantify the pruning impact on the model robustness, namely \emph{$L_1$-norm} and \emph{entropy}.
The $L_1$-norm measures the overall scale of pruning impact on the model's output, in a way that a smaller value tends to incur less uncertainty in the network's decision making.
The entropy measures the similarity of the pruning impact on each output unit. A smaller entropy is obtained in the scenario that the pruning impact is more uniformly distributed in each output unit, and therefore implies that the pruned model is less sensitive when dealing with undesired perturbations in inputs.

We implement our supervised data-free pruning method in a Python package and evaluate it with a series of experiments on diverse neural network models.
The experimental results show that our supervised pruning method  offers promising fidelity and robustness preservation. All the tested models preserves at least 50\% of its original accuracy after 50\% of hidden units have been pruned, and 50\% of its original robustness even after 70\% of hidden units have been pruned.
It significantly outperforms existing one-shot data-free approaches in terms of both robustness preservation and accuracy, with improvements up to 42\% and 66\%, respectively.
Our evaluation also demonstrates that it can generalize on a wide range of neural network architectures, including the fully-connected (FC) multilayered perceptron (MLP) models and convolutional neural network (CNN) models.

In summary, the contributions of this work are as follows.
\begin{itemize}[noitemsep,topsep=2pt,leftmargin=1.3em]
	\item \textbf{A robustness-preserving data-free pruning framework.}
	We investigate the robustness-preserving neural network pruning in the data-free context.
	To the best of our knowledge, this is the first work of this kind.
	
	\item \textbf{A stochastic pruning method.}
	We reduce the pruning problem into a stochastic process, to replace the coarse-grained one-shot pruning strategy.
	The stochastic pruning is solved with the simulated annealing algorithm.
	This avoids mis-cutting off those hidden units that play crucial roles in the neural network's decision making.
	
	\item \textbf{Implementation and evaluation.}
	We implement our punning method into a Python package and evaluate it with a series of experiments on representative datasets and models.
	Our evaluation covers not only those models trained on datasets commonly used in the research community such as MNIST and CIFAR-10, but also models designed to solve real-world problems such as credit card fraud detection and network intrusion, demonstrating that our proposed pruning can generalize on different robustness-sensitive tasks.
\end{itemize}

We have made our source code available online\footnote{\url{https://github.com/mark-h-meng/nnprune}} to facilitate future research on the model pruning area.


%% file: sections/background.tex
\subsection{Neural Network Pruning}
A typical deep neural network is a MLP architecture that contains multiple fully connected layers.
For this reason, deep neural networks are widely recognized as an over-parameterized and computationally intensive machine learning technique~\cite{ba2014deep}.
Neural network pruning was introduced as an effective relief to the performance demand of running them with a limited computational budget~\cite{lecun1990optimal}. In recent years, as deep neural networks are increasingly applied in dealing with complex tasks such as image recognition and natural language processing, network pruning and quantization are identified as two key model compression techniques and have been widely studied~\cite{choi2020universal,tang2020scop,liu2019rethinking,han2015learning,he2017channel,yu2018nisp,chin2018layer,hu2016network,li2016pruning,suau2020filter}.
Existing pruning techniques could be grouped into two genres. One genre of pruning is done by selectively zeroing out weight parameters (also known as synapses).
This type does not really reduce the size and computational scale of a neural network model, but only increases the sparsity (i.e., the density of zero parameters)~\cite{wen2016learning}.
Therefore, that genre of pruning is categorized as \emph{unstructured pruning} in the literature~\cite{liu2019rethinking,han2015learning}.
In contrast, the other genre called \emph{structured pruning} emphasizes cutting of entire hidden unit with all its synapses off from which layer it is located, or removal of specific channel or filter from a convolutional layer~\cite{luo2017thinet,srinivas2015data,tang2020scop}.

\emph{Pruning target} is the common metric to assess neural network pruning.
It indicates the percentage of parameters or hidden units to be removed during the pruning process, and therefore it is also known as \emph{sparsity} in some literature on unstructured pruning.
\emph{Fidelity} is another metric that describes how well the pruned model mimics the behavior of its original status and is usually calculated through accuracy.
An ideal pruning algorithm with promising fidelity should not incur a significant accuracy decline when compared with the original model.
However, the discussion of the impact of pruning to measurements beyond fidelity, such as robustness, is still in its nascent phase~\cite{liebenwein2021lost}.
As robustness is a representative property specification of a neural network model that concerns the security of its actual deployment, unveiling the influence of pruning on robustness could provide a guarantee to the trustworthiness of pruning techniques.

\vspace{-0.1cm}
\subsection{Stochastic Optimization}

Stochastic optimization refers to solving an optimization problem when randomness is present. In recent years, stochastic optimization has been increasingly used in solving software engineering problems such as testing~\cite{su2017guided} and debugging~\cite{le2015finding}.
The stochastic process offers an efficient way to find the optimum in a dynamic system when it is too complex for traditional deterministic algorithms.
The core of stochastic optimization is the probabilistic decision of its transition function in determining whether and how the system moves to the next state.
Due to the presence of randomness, stochastic optimization has an advantage in escaping a \emph{local optimum} and eventually approaching the \emph{global optimum}.

The \emph{simulated annealing algorithm}~\cite{van1987simulated} is an extensively used method for stochastic optimization.
It is essentially proposed as a \emph{Monte Carlo} method that adapts the Metropolis-Hastings algorithm~\cite{chib1995understanding} in generating new states of a thermodynamic system.
At each step, the simulated annealing calculates an \emph{acceptance rate} based on the current temperature, generates a random probability, and then makes a decision based on these two variables.
In case the generated probability is less than the acceptance rate, the system accepts the currently available neighboring state and accordingly moves to the next state; otherwise, it stays at the current step and then considers the next available neighboring candidate.
In general, the simulated annealing algorithm provides an efficient approach to drawing samples from a complex distribution.

%% file: sections/problem.tex

\subsection{Robustness of Neural Networks}

Robustness is a feature representing the \emph{trustworthiness} of a neural network model against real-world inputs.
The real-world inputs may be from an undesired distribution, and are often with distortions or perturbations, either intentionally (e.g., adversarial perturbations) or unintentionally (e.g., blur, weather condition, and signal noise)~\cite{meng2022adversarial}.
For this reason, robustness is particularly crucial in the open-world deployment of a neural network model.


The evaluation of robustness is discussed against adversarial models, such as projected gradient descent (PGD) attack and fast-gradient sign method (FGSM).
Take FGSM as an example, the adversary can generate an $L_{\infty}$-norm \emph{untargeted perturbation} for an arbitrary test sample.
The untargeted perturbation is calculated with the negative sign of the loss function's gradient and then multiplied with $\epsilon$ before adding to the benign input. The $\epsilon$ is usually a very small fraction to ensure the adversarial samples are visually indistinguishable from those benign ones~\cite{goodfellow2015explaining}.
By doing that, an adversarial input tends to maximize the loss function of the victim neural network model and thereby leads the model to misclassify.
The FGSM is an effective attack model to evaluate robustness and has been extensively applied in both literature~\cite{bastani2016measuring,goodfellow2015explaining} and mainstream toolkits such as TensorFlow. Accordingly, we adopt FGSM as the default attack model and assume input perturbations are measured in $L_{\infty}$-norm in this work.

Given an adversarial strategy, the attacker can modify an arbitrary benign input with a crafted perturbation to produce an adversarial input.
We formalize the robustness property of the neural network model as follows.

\begin{definition}[Robustness against adversarial perturbations] \label{def_robustness}
	Given a neural network model $f$, an arbitrary benign instance $x$ sampled from the input distribution (e.g., a dataset) $X$, and an adversarial input $x_{\text{adv}}$ which is produced by a specific adversarial strategy based on $x$, written as $x_{\text{adv}}=adv\left(x\right)$. The model $f$ satisfies the robustness property with respect to $x$, if it makes consistent predictions on both $x$ and $x_{\text{adv}}$, i.e., $ f\left( x \right) =f\left( x_{\text{adv}} \right) $.
\end{definition}

\subsection{Robustness-preserving Pruning}

Our pruning method aims to preserve the robustness of a given neural network model.
Following a previous study~\cite{bastani2016measuring}, we define this preservation as the extent that the pruned model can obtain the maximum number of consistent predictions of both benign and adversarial inputs, and accordingly, we name those inputs of consistent predictions as \emph{robust instances}.
Thus, we propose an objective function specifying the number of robust instances from a given distribution.
The robustness-preserving pruning then becomes an optimization problem that aims to identify a pruning strategy towards maximizing the objective function.
We formalize our goal of robustness-preserving pruning as follows.

\begin{definition}[Robustness-preserving pruning] \label{def_goal}
	Given a neural network model $f$ that takes inputs and labels from distribution $X$. Each input $x$ has a corresponding label $y$, written as $\left(x,y\right)\in X$. Let $x_{\text{adv}}$ be the adversarial input that adds perturbations to a benign input $x$. Our goal is to find a pruning method $\pi$ that transforms the original neural network model $f$ to a pruned one $g$, which \emph{maximizes} the objective function $\mathbb{Z}(\pi)$ that counts the occurrence of robust input instances from the distribution $X$, written as:
	\begin{equation}
	\mathbb{Z}(\pi) = \left| \left\{ x| \left(g\left( x_{\text{adv}} \right) =g\left( x \right)=y\right)  \land \left(x,y\right) \in X \right\} \right|
	\end{equation}
\end{definition}

%% file: sections/overview.tex

\begin{figure}[t!]
	\centering
	\includegraphics[trim={0 0cm 0 0cm},clip,width=0.95\linewidth]{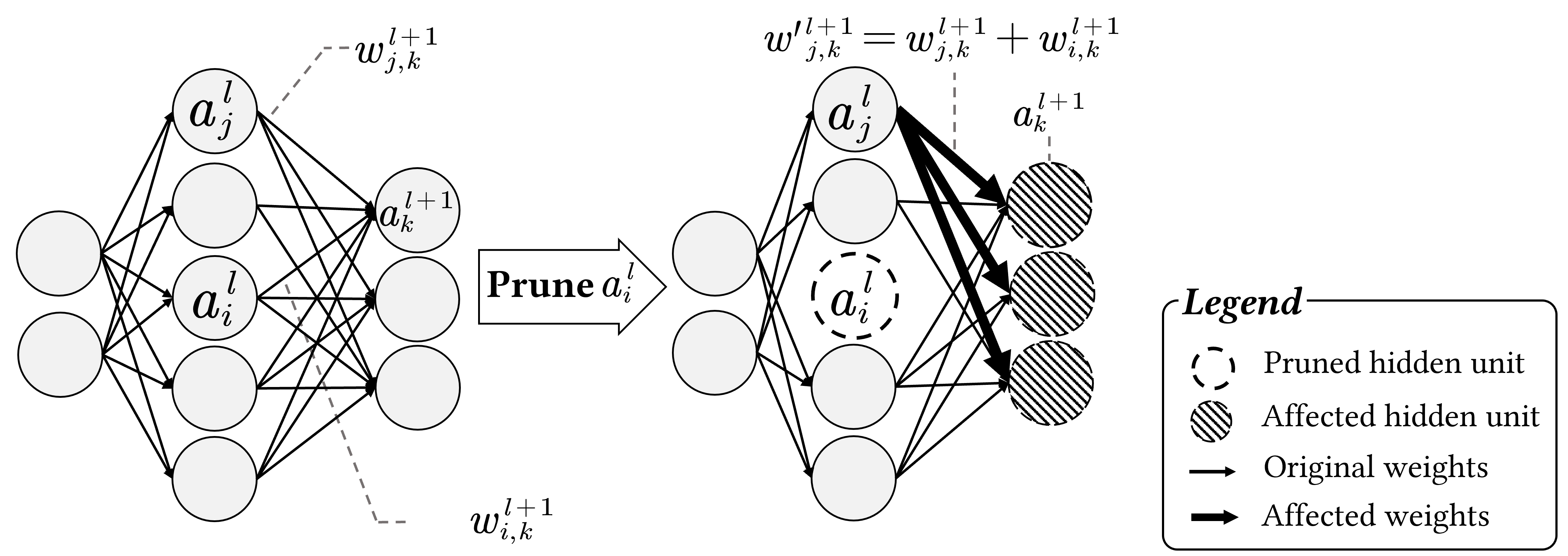}
	\vspace{-2pt}
	\centering
	\captionsetup{justification=centering}
	\caption{An illustration of the primitive pruning on $\left<a^l_i, a^l_j\right>$}
	\vspace{-10pt}
	\label{fig:pruning}
\end{figure}

\subsection{Saliency-based Primitive Pruning Operation}
%
When attempting to prune a hidden unit~(denoted by the \emph{nominee}, i.e., the one chosen to be pruned), our method uses a pair-wise strategy rather than simply deleting the nominee.
In particular, our primitive pruning operation considers another unit~(denoted by the \emph{delegate}, i.e., the one to cover the nominee's duty) from the nominee's layer that tends to play a similar role in making a prediction.
It removes the nominee and adjusts the parameters of the delegate so that the impact of a single pruning operation on the subsequent layers can be reduced.
Given a nominee and delegate pair $\left<a^{l}_{i},a^{l}_{j}\right>$, which are the $i$-th and $j$-th hidden units at the layer $l$, the primitive pruning operation performs the following two steps.

\vspace{5pt}
\begin{enumerate}[noitemsep,topsep=0pt]
\addtolength{\itemindent}{0.75cm}
\item[\textbf{Step (1)}] The nominee $a^{l}_{i}$ is pruned. To this end, we zero out all parameters connecting from and to $a^{l}_{i}$;
\item[\textbf{Step (2)}] We modify the parameters connecting from the delegate $a^{l}_{j}$ to the next layer with the \emph{sum} of the parameters of both $a^{l}_{i}$ and $a^{l}_{j}$.
\end{enumerate}
\vspace{5pt}

The parameter update in Step (2) is carried out to offset the impact caused by pruning the nominee. Fig.~\ref{fig:pruning} illustrates our primitive pruning operation.

To find the delegate, we use a metric called \emph{saliency}, which is proposed in a previous study~\cite{srinivas2015data} to assess the ``importance'' of a unit when it is to be replaced by another unit in its layer.
A lower saliency means that the nominee can be replaced by the delegate with less impact on the network.
Let $w^l_{i,j}$ be the weight parameter connecting the $i$-th hidden unit at the layer $l-1$ with the $j$-th hidden unit at the layer $l$, and $b^l_i$ be the bias parameter of the $i$-th hidden unit at the layer $l$.
Given the nominee $a^l_i$, its saliency with respect to the delegate $a^l_j$ is measured as follows.
\begin{multline}\label{eq:SAL}
	S\left( a^l_i,a^l_j \right) = \frac{\sum{W^{l+1}_{i,*}}}{\left| W^{l+1}_{i,*} \right|} \left( \lVert W^l_{*,i}-W^l_{*,j} \rVert _2+\frac{\left| b^l_i-b^l_j \right|}{\left| b^l_i+b^l_j \right|} \right), \\
	\qquad\qquad\qquad \text{where} \; W_{i,*}^{l}=\left\{ w_{i,m}^{l}|a_{m}^{l}\in a^{\left[ l \right]} \right\},\\
	W_{*,i}^{l}=\left\{ w_{n,i}^{l}|a_{n}^{l}\in a^{\left[ l-1 \right]} \right\}
\end{multline}

\begin{figure}[t!]
	\centering
	\includegraphics[trim={0 0 0 0},clip,width=1\linewidth]{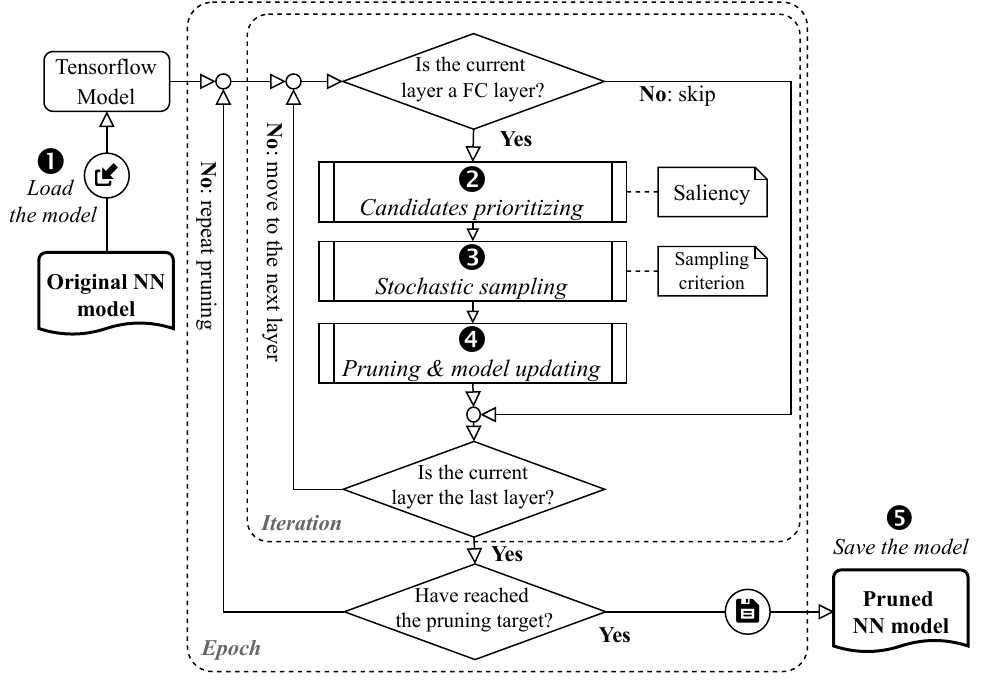}
	\centering
	\captionsetup{justification=centering,margin=0cm}
	\caption{The workflow of our pruning method}
	\label{fig:overview}
	\vspace{-15pt}
\end{figure}

\subsection{Workflow of the Pruning Method}


Fig.~\ref{fig:overview} shows the workflow of our pruning method.
It begins with reading a pre-trained model and loading its architecture and parameters (Stage~\ding{182}).
Then it traverses the model layer by layer and iteratively performs hidden unit pruning.
The pruning process (Stage \ding{183}-\ding{185}) might be executed in multiple epochs, depending on the pruning target and pruning batch size per epoch.
Once the pruning target has been reached, our method saves the pruned model (Stage~\ding{186}).

Below we brief each component in the pruning process.
The outer loop specifies an epoch, in which a fixed portion of fully connected units (i.e., the batch size) will be cut off from the model.
The inner loop represents the iteration of all layers in a forward direction.
A pruning iteration is composed by three stages, i.e., \emph{candidates prioritizing} (Stage \ding{183}), \emph{stochastic sampling} (Stage \ding{184}) and \emph{pruning and model updating} (Stage \ding{185}).
The former two stages identify the units to be pruned at each iteration, and then our method invokes the primitive pruning operation to prune each of them.

\begin{itemize}[noitemsep,topsep=2pt,leftmargin=1.5em]
  \item The candidates prioritizing stage evaluates the saliency for every pair of hidden units at the beginning of each iteration and generates a saliency matrix.
  Considering that pruning a nominee that is hard to find a proper delegate~(i.e., the nominee has high saliency with respect to every hidden unit in its layer) is unfavorable,
  we sort the list of hidden unit pairs according to their saliency values in ascending order and pass that list to the next stage.
  Those candidates with the least saliency values are given priority to be processed in the next stage.

  \item The stochastic sampling stage takes the list of pruning candidates as input and identifies the units to be pruned.
  The basic idea is to estimate how pruning a unit impacts the prediction at the output layer, to decide whether to keep or discard it.
  A naive way is to evaluate the impact of each candidate, but this is too costly since calculating each impact requires a forward propagation till the output layer.
  We thus employ a stochastic sampling strategy with the estimated impact as a guide in this process.
  Our impact estimation and corresponding sampling strategy are detailed in Section~\ref{sec:methodology}.

\end{itemize}

%% file: sections/stochastic_sampling.tex

In this section, we introduce our supervised pruning method.
We detail our approach of estimating how pruning a unit impacts the prediction at the output layer~(Section~\ref{sec:impact}).
With this, we can approximate the cumulative impact on the robustness of the final model, and thus we embed it into our sampling criterion~(Section~\ref{sec:criterion}).
To prevent the sampling method from being stuck at a local optimum, we employ the simulated annealing algorithm in determining which candidate(s) to prune~(Section~\ref{subsec:guided-pruning}).

\subsection{Estimation of Pruning Impact} \label{sec:impact}

Because the primitive pruning operation prunes the nominee and modifies the value of parameters connecting from the delegate to the next layer,
it affects the computation of the hidden units in subsequent layers.
Such impact would eventually propagate to the output layer.
In this section, we discuss how we estimate this impact.

For an original model $f$ that performs an $n$-class classification, its output, when given a sample $x$, is a vector of $n$ numbers denoted by $f\left(x\right)=\left[o_1,...,o_n\right]$.
The model $g$, which is derived by pruning a unit of $f$, outputs another vector of the same size, denoted by $g\left(x\right)=\left[o'_1,...,o'_n\right]$.
We aim to estimate the impact at the output layer as a vector of $n$ items, i.e., $g\left(x\right)-f\left(x\right)$ for any legitimate input.
To achieve this, we first approximate the valuation of hidden units involved in the candidate pair~(i.e., nominee and delegate) by interval arithmetic based on the bounds of normalized input.
Next, we assess the impact caused by a primitive pruning operation on the subsequent layer where the pruning operation is performed.
In this process, we quantify the impact as an interval.
After obtaining the impact on the subsequent layer, we apply the forward propagation until the output layer, so that the pruning impact on the output layer can be derived.

We adopt the abstract interpretation that is commonly used in the literature of neural network verification~\cite{wang2018efficient,singh2019abstract} to estimate the upper and lower bounds of an arbitrary hidden unit.
To achieve that, we need to define the scope of a legitimate input as an interval. As input normalization is a common pre-processing practice prior to model training, the value of an input feature is usually restricted to a fixed range (e.g., $\left[0,1\right]$).
With a vector of intervals provided as the input, we perform the forward propagation to approximate the valuation of the involved hidden units.
This propagation simulates the computation within a neural network model with a specific input.
During the propagation, we leverage the interval arithmetic to calculate the upper and lower bounds.
In the actual implementation, we build a map of intervals for all hidden units of the neural network at the beginning of pruning.
Due to each primitive pruning modifies parameters at the next layer, we update the map with the latest estimation after each iteration specifies the batch pruning at the same layer.

For an arbitrary hidden unit at the $\left(l+1\right)$-th layer $a^{l+1}_k\in a^{\left[l+1\right]}$, its impact caused by a primitive pruning of $\left<a^l_i, a^l_j\right>$ can be formulated as Eq.\ref{eq:delta_y} below.
\begin{equation}\label{eq:delta_y}
	\begin{split}
		\Delta a_{k}^{l+1} & =
		\left( w^{l+1}_{j,k}+w^{l+1}_{i,k} \right) a^{l}_{j} -\left( w^{l+1}_{i,k} a^{l}_i + w^{l+1}_{j,k} a^{l}_{j}\right) \\	
		& = w^{l+1}_{i,k} \left( a^{l}_{j}-a^{l}_{i} \right)
	\end{split}
\end{equation}
We can obtain the latest estimation of both $a^l_i$ and $a^l_j$ from the map of intervals that we have built at the beginning. Since all weight parameters are known in the white-box setting, we can also quantify the impact $\Delta a_{k}^{l+1}$ as an interval.

Next, we perform another round of forward propagation to simulate the impact of affected hidden units from the layer $l+1$ to the output layer. The value to be propagated in this round is no longer the interval of input, but the impact of affected hidden units as intervals. The propagated impact at the output layer could be treated as the estimated result of $g\left(x\right)-f\left(x\right)$ for the current pruning operation, where $f$ and $g$ stand for the original model and the pruned model, respectively. 
The propagated impact on the output for each pruning operation will be accumulated along with the pruning progress. We call it \emph{cumulative impact} to the output layer and use $\Delta a^{\left[out\right]}$ to denote it in the remaining of this section.

\subsection{Sampling Criterion} \label{sec:criterion}

Our sampling criterion is proposed based on an insight that a small and uniformly distributed cumulative impact is less possible to drive the pruned model to generate an output that is different from the original one, even the input is with an adversarial perturbation.
On the contrary, the pruning impact with a variety of scales and values is considered to impair the robustness because it makes the pruned model sensitive that its prediction may flip when encountered a perturbation in the input.
Our proposed criterion is composed of two metrics.

\begin{itemize}[noitemsep,topsep=2pt,leftmargin=1.5em]
  \item One metric accounts for the \emph{scale} of cumulative impact on the output layer. A greater scale means the current pruning operation generates a larger magnitude of impact on the output layer.
  \item The other is based on the \emph{entropy} that assesses the degree of similarity of cumulative impact on each output unit. A greater entropy implies the pruning impact on each output node shows a lower similarity.
\end{itemize}

Our sampling strategy jointly considers both metrics and favors both to be small.

\subsubsection*{\textbf{Metric \#1: Scale}} \label{metric1}

As we can obtain the cumulative impact as a vector of intervals, we adopt the $L_1$-norm to assess the scale of cumulative impact.
Here we use $\left( u^-, u^+ \right)$ to represent the impact bounds of an arbitrary node $u$ at the output layer and let the term $NORM$ denote the $L_1$-norm of the intervals. The formula to calculate $NORM$ is shown in Eq.~\ref{eq:NORM}.

\begin{equation}\label{eq:NORM}
	NORM\left( \Delta a^{\left[out\right]}\right) = \sum_{\left( u^-, u^+ \right) \in \Delta a^{\left[out\right]}}{\left| u^+-u^- \right|}
\end{equation}

\subsubsection*{\textbf{Metric \#2: Entropy}} \label{metric2}
We apply \emph{Shannon's information entropy}~\cite{shannon1948mathematical} to measure the similarity of cumulative impact on each output unit.
Our measurement of the {similarity} for a pair of intervals is adopted from existing literature~\cite{zhang2006similarity,dai2016attribute}, as defined below.

\begin{definition}[Similarity of interval-valued data] \label{def_similarity}
	Given a list of intervals $U=\left\{u_1, u_2, ..., u_3\right\}$, and each interval is composed of its lower and upper bounds such as $u_i=\left[u_i^{-}, u_i^{+}\right]$. Let $m^{-}=\min _{u_i\in U}\left\{ u_i^{-} \right\}$ be the global minimum of $U$, i.e., the minimum of lower bounds, and similarly, $m^{+}=\max _{u_i\in U}\left\{ u_i^{+} \right\}$ be the global maximum. The similarity degree of relative bound difference between two intervals $u_i$ and $u_j$ is defined as:
\begin{equation}\label{eq:similarity}
	Sim_{ij}=1-\frac{1}{2}\frac{\left| u_i^{-}-u_j^{-} \right|+\left| u_i^{+}-u_j^{+} \right|}{m^{+}-m^{-}}
\end{equation}	
	With this definition, we say two intervals $u_i$ and $u_j$ are $\phi$-similar if $Sim_{ij} \ge \phi$ for any similarity threshold $\phi = \left[0, 1\right]$.
\end{definition}

Next, we measure the overall similarity of an interval with all other intervals in a list, and we call it \emph{density of similarity}.
We adopt the calculation of the density of $\phi$-similarity for an interval from an existing study~\cite{dai2016attribute}, which is defined as follows.

\begin{definition}[Density of similarity] \label{def_similarity_density}
	For an interval $u_i$ from a list of intervals $U=\left\{u_1, u_2, ..., u_3\right\}$, its density of $\phi$-similarity among $U$ is measured by the probability of an arbitrary interval (other than $u_i$) is $\phi$-similar with itself, calculated as:
	\begin{equation}
	\rho _{\phi}\left( u_i \right) =\frac{\left| \left\{ u_j|Sim_{ij}\ge \phi ,\forall u_j\in U\backslash \left\{ u_i \right\} \right\} \right|}{\left| U \right|}
	\end{equation}
\end{definition}

With the density of similarity, we define the metric as the \emph{entropy} of the cumulative impact on the output layer, written as $ENT$. The formula of $ENT$ calculation is shown in Eq.\ref{eq:ENT}.

\begin{equation}\label{eq:ENT}
	ENT\left( \Delta a^{\left[out\right]}\right) = -\sum_{u_i \in \Delta a^{\left[out\right]}}{\rho _{\phi}\left( u_i \right) \cdot \log \rho _{\phi}\left( u_i \right) }
\end{equation}

The similarity threshold $\phi$ is in the range $\left[0,1\right]$.
With the same set of intervals, a higher $\phi$ results in a lower density of similarity such that it makes entropy calculation more sensitive to the difference of those intervals.
In our work, the cumulative impact is obtained after a forward propagation of several layers and therefore might be in a large magnitude.
Accordingly, we set 0.9 as the default value of $\phi$ to maintain a variety of similarity densities rather than all equal to one\footnote{A comparably greater value of $\phi$ is needed to maintain a favorable distinguishable degree among hidden units' outputs rather than always producing a similarity density equals 1. For this reason, the value 0.9 is used. }.

\subsubsection*{\textbf{Synthesis}} \label{joint}
We introduce a pair of parameters $(\alpha, \beta)$ to specify the weight of these two metrics.
Considering these two metrics may have different magnitude, and particularly, the $NORM$ is unbounded (i.e., no upper bound), we use a sigmoid function ($\sigma$) to normalize these two metrics in the final criterion.
Due to the concave and monotonic nature of sigmoid logistic function for values greater than zero, it can output bounded results within $\left(-1,1\right)$ with their values' order the same with input, denoted as $x_1 < x_2 \Leftrightarrow \sigma\left(x_1\right) < \sigma\left(x_2\right)$.
On the whole, the definition of our sampling criterion is given in Eq.~\ref{eq:final-criterion} below.


\begin{equation}\label{eq:final-criterion}
	\begin{split}
	Energy_{\left(\alpha,\beta\right)}\left(\Delta a^{\left[out\right]}\right) = \alpha \cdot \sigma\left(NORM\left(\Delta a^{\left[out\right]}\right)\right) \\
		+ \beta \cdot \sigma\left(ENT\left(\Delta a^{\left[out\right]}\right)\right), \qquad & \\
	\text{subject to} \; \alpha \ge 0, \beta \ge 0, \beta = 1-\alpha &
	\end{split}
\end{equation}

We use the term \emph{energy} to represent our sampling criterion to echo the simulated annealing algorithm used in our guided stochastic sampling strategy, which will be presented in the next subsection.

\subsection{Guided Stochastic Sampling}

\label{subsec:guided-pruning}

Since the sampling criterion can reflect the impact of the unit pruning on the model robustness, a naive way is to calculate $energy$ (Eq.~\ref{eq:final-criterion}) for every pair and prune the unit with the least value.
However, this is too expensive because each calculation requires a forward propagation in a fully connected manner.
To address this, we use a stochastic sampling guided by the $energy$-based heuristic to identify the candidates to be pruned.
Our method is presented in Algorithm \ref{guided_pruning_alg}, and we discuss it in the remainder of this section.

\input{sections/algorithm}


We exploit the idea of \emph{simulated annealing} to implement our sampling strategy through the lens of stochastic optimization.
In particular, our method traverses the hidden unit pairs from the candidates prioritizing result one by one.
In the beginning, our method by default accepts the first candidate from the prioritizing result and records its $energy$ as the evaluation of the current state.
Upon receiving a new pruning candidate, the method calculates the $energy$ of that candidate, compares it with the current state, and decides whether to prune it during the current iteration, according to an acceptance rate calculated based on a temperature variable $T$.
The temperature variable is adopted from the thermodynamic model. The descent of temperature value reflects the \emph{solving progress} of the optimization problem -- as temperature decreases, our method would less possibly accept a pruning candidate with an $energy$ greater than the current state. Here we define the temperature used in our method as the portion of the remaining pruning task, which equals 1 at first and approaches 0 when the pruning target is reached.
Given the temperature of the current iteration written as $T_{t-1}$, the assessment of the last drawn (accepted) candidate $energy_{t-1}$, we can obtain the acceptance rate of the next candidate (line 12 of Algorithm~\ref{guided_pruning_alg}) once we calculate its energy (written as $energy'$, line 10) according to Eq.~\ref{eq:final-criterion}.
The formula of acceptance rate is provided as follows.
\begin{equation}\label{eq:acceptance_rate}
	P=\min \left( 1,\exp \left( -\frac{energy'-energy_{t-1}}{T_{t-1}} \right) \right)
\end{equation}
As Eq.~\ref{eq:acceptance_rate} shows, our method automatically accepts a candidate if its $energy$ is lower than the one in the current state; otherwise, a random probability will be generated and tested against the acceptance rate to determine whether we accept or discard the candidate. This procedure is reflected as lines 11-20 of Algorithm~\ref{guided_pruning_alg}.

There are two obvious advantages of this stochastic process. First, applying such randomization in sampling is less expensive than computing $energy$ of all candidates and sorting.
Moreover, the stochastic process through simulated annealing enables us to probabilistically accept a candidate that may not have the lowest $energy$ at the current step.
This helps prevent our pruning method from being stuck at a local optimum and eventually achieves our objective.

%% file: sections/algorithm.tex
\begin{algorithm}[t]
	\small
	\caption{Supervised pruning with a stochastic heuristic\label{guided_pruning_alg}}
	\begin{algorithmic}[1]
		\renewcommand{\algorithmicrequire}{\textbf{Input:}}
		\renewcommand{\algorithmicensure}{\textbf{Output:}}
		\REQUIRE An $n$-layer model to be pruned ($q_{t-1}$), cumulative impact of all previous pruning ($\Delta a^{\left[out\right]}$), weights used in sampling criterion ($\alpha, \beta$), batch size ($k$), current temperature $T_{t-1}$
		\ENSURE  A pruned deep learning model $q_t$, updated cumulative impact of pruning $\Delta a^{\left[out\right]}$, the list of hidden units pruned $p^l$, the updated temperature $T_t$
		\vspace{3pt}
		
		\FOR {layer $l$ in all hidden layers}
		\STATE load parameters of the current layer $\Sigma_{l}$
		\STATE build a saliency matrix $M^l$ for the unit pairs
		\STATE sort the saliency matrix $M^l$ in ascending order
		\STATE set $energy_{t-1} \gets 0$
		\FOR {hidden unit pair $\left<i,j\right>$ in the first $k$ values in $M^l$}
			\STATE simulate a pruning of $\left<i,j\right>$ and calculate the impact on the output layer $\Delta _{p}$
			\STATE calculate temporary cumulative impact $temp\Delta a^{\left[out\right]} \gets \Delta a^{\left[out\right]} + \Delta_{\left<i,j\right>}$
			\STATE calculate $NORM$ and $ENT$ with $temp\Delta a^{\left[out\right]}$
			\STATE calculate sampling criterion $energy' \gets \alpha \cdot \sigma\left(NORM\right) + \beta \cdot \sigma\left(ENT\right)$
			\IF {($energy_{t-1} > 0$) \textbf{and} ($energy' > energy_{t-1}$)}
				\STATE calculate acceptance rate $P$ based on temperature $T_{t-1}$ and $energy'$
				\STATE generate a random probability $rand$
				\IF {($rand \ge P$)}
					\STATE \textbf{\textit{reject}} the current $\left<i,j\right>$ and go to the next one
				\ENDIF
			\ENDIF
			\STATE append $\left<i,j\right>$ into a pruning candidate list $\mathcal{L}^{l}$
			\STATE \textbf{\textit{accept}} and perform pruning 
			\STATE update $energy_{t-1} \gets energy'$
		\ENDFOR
		\STATE update $\Delta a^{\left[out\right]}$ with $\mathcal{L}^{l}$
		\STATE update temperature $T_t \leftarrow T_{t-1}$
		\ENDFOR
	\end{algorithmic} 
\end{algorithm}

%% file: sections/experiment1.tex
This section presents the evaluation of our pruning method.
We aim to answer the following three research questions.

\begin{itemize}[noitemsep,topsep=2pt,leftmargin=1.5em]
  \item \textbf{RQ1: Fidelity and Robustness Preservation.} How effective is our pruning method in terms of fidelity and robustness preservation? Does our method generalize on diverse neural network models?
  \item \textbf{RQ2: Pruning Efficiency.} Can our method complete the pruning within an acceptable time?
  \item \textbf{RQ3: Benchmarking.}  Can our method outperform one-shot strategies in terms of robustness preservation?
\end{itemize}

%
\subsection{Implementation and Experiment Settings}
We implement our pruning method in Python.
All neural network models are trained, pruned, and evaluated based on TensorFlow. 
Our toolkit accepts any legitimate format of neural network models trained by TensorFlow.
It also allows the user to configure the pruning target and the number of pruning per epoch.
Given a model as input, it automatically identifies the fully-connected hidden layers, prunes the hidden units from them, and stops once the pruning has reached the setting threshold (e.g., 80\% of hidden units have been cut off).

To evaluate our method on diverse mainstream neural network applications, we select four representative datasets ranging from structured tabular data to images, with labels for both \emph{binary classification} and \emph{multi-class classification}.
For each dataset, we select a unique architecture of neural network model to fit the classification task.
To this end, we refer to the most popular example FC or CNN models on \emph{Kaggle}\footnote{https://www.kaggle.com/ (accessed in July 2022). }.
We have trained four models with different architectures, covering both purely fully connected MLPs and CNNs.
All models are equipped with ReLU activation and are trained with a 0.001 learning rate for 20 epochs.
The number of fully connected layers and hidden units per layer varies among these models.
The diversification of tested models is to evaluate the generalizability of our method~(\textbf{RQ1}).
The details of the four used datasets and pre-trained models are listed in Table~\ref{tab:table-datasets}.

\input{sections/table_of_dataset_and_models_single_col}
 
We empirically select the values of the parameters $\alpha$ and $\beta$ through a tuning process.
We observe the pruning of those multi-class models like MNIST and CIFAR-10 is more sensitive to their values compared with binary classification models.
As the result, we find that $\alpha=0.75$ and $\beta=0.25$ suits all the tested models, because these models use ReLU activation where the $L_1$-norm of pruning impact plays a more important role than entropy.
Our experiments run on an 8-core Intel CPU (2.9GHz Core (TM) i7-10700F), 64GB RAM and an NVIDIA GeForce RTX 3060Ti GPU.

\begin{figure}[t!]
	\centering
	\includegraphics[trim={0 0 0 0cm},clip,width=1\linewidth]{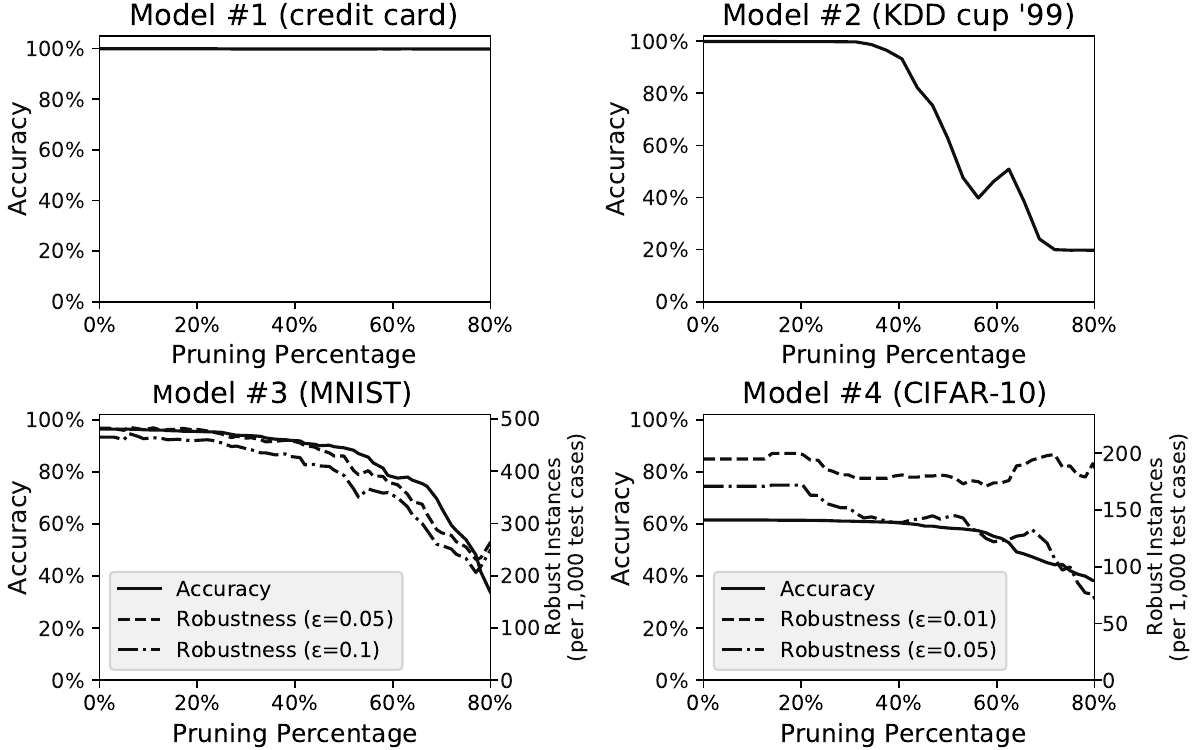}
	\caption{Accuracy and robustness decay of four models when applying our supervised pruning method (up to 80\% pruning)}
	\label{fig:decay}
	\vspace{-15pt}
\end{figure}

\subsection{RQ1: Fidelity and Robustness Preservation}
Our first set of experiments is conducted to investigate the fidelity and robustness preservation of our pruning method.
As previous studies have shown that pruning may cause the loss of model fidelity~\cite{srinivas2015data,liu2019rethinking},
we test the accuracy of all four models during the pruning to explore the impact of our method on it.
This is crucial because a poor accuracy would undermine the validity of robustness which only requires the model not to produce inconsistent outputs for a given benign input and its adversarial variant, regardless of whether the benign input is correctly predicted or not.

We also test the robustness preservation based on the metric given in Definition~\ref{def_goal}.
Our evaluation calculates the number of consistent and correct classification of adversarial inputs on both the original and pruned models, and compares these two numbers to determine the degree of robustness preservation.
We apply this method on two models (\#3-4) because their datasets, MNIST and CIFAR-10, have been widely studied on the subject of model robustness and trustworthiness.

With the overall pruning target set as 80\% of hidden units being pruned, our method prunes the same proportion of units per layer at each iteration.
Both the accuracy and robustness, if applicable, are evaluated after each pruning epoch.

Fig.~\ref{fig:decay} shows the change of accuracy for each model during the pruning process. It also tracks the robustness preservation of our method on models \#3-4 against untargeted FGSM adversaries with different epsilon ($\epsilon$) options.
This parameter is used in FGSM to measure the variation between the adversarial and benign samples. We refer to the literature~\cite{bastani2016measuring,goodfellow2015explaining} to find proper values to use in our experiment\footnote{We select two values ($\epsilon=0.01$ and $0.05$) for models \#3-4 to examine their robustness against perturbations in different sizes.}.
We perform 10 rounds of experiments on all four models and plot the median of the experimental results in the figure.

In general, our method performs well on all four models.
On the model for binary classification, i.e., model \#1, our method imposes almost no impact on the robustness and accuracy, even when 80\% of units are pruned.
For those models with more complex classification tasks, i.e., models \#2-4,
our method still achieves favorable results.
All models preserve 50\% of their original accuracy when 50\% of their units have been pruned.
Both models \#3 and \#4 preserve at least 50\% of their original robustness even after 70\% of hidden units are pruned.
The change of robustness generally shares the same trend as test accuracy (see Fig.~\ref{fig:decay}).

In models \#3-4, we observe that the robustness slightly grows as the number of pruned units increases.
This is because these models are not trained with robustness preservation as part of the objective functions, and
our pruning guided by that metrics which incorporate robustness preservation may enhance their robustness.
This on the other hand demonstrates the effectiveness of the robustness preservation of our method.

Our first set of experiments has responded to \textbf{RQ1}.
In summary, our pruning method shows favorable fidelity and robustness preservation against adversarial perturbations.
It does not show any drastic performance decay along with the pruning progress.
Our method is also able to generalize on many types of models, as shown by the evaluation outcomes of the four tested models.

\input{sections/table_of_elapsed_time}

\subsection{RQ2: Pruning Efficiency}
To explore the efficiency of our method, we run it on the four models with a ``worst-case setting''.
Specifically, we examine the case of pruning a large proportion~(80\%) of the entire model, in the slowest pace~(1 or 2 per layer per epoch).
This is not necessary at all and gives our method disadvantages, but when applied in practice, it could be much more efficient.

Table~\ref{tab:table-elapsed-time} details the time consumed by our method on each of the four models.
In general, our method can prune a model within an acceptable time.
In the multi-class prediction models, the pruning process can be completed within 8 minutes,
while in the binary classification models, the process can be completed much faster.

\begin{figure}[t!]
	\centering
	\includegraphics[trim={0 0cm 0 0cm},clip,width=1\linewidth]{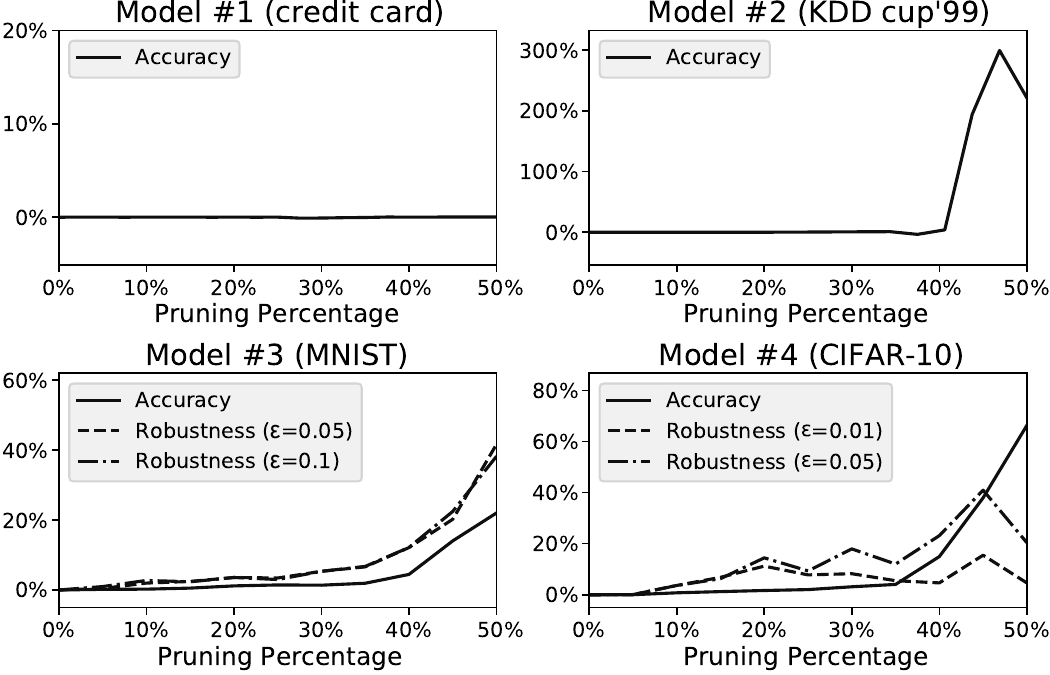}
	\centering
	\captionsetup{justification=centering}
	\caption{Improvement of our method against saliency-based one-shot pruning on four models}
	\label{fig:improvement}
	\vspace{-15pt}
\end{figure}

\subsection{RQ3: Benchmarking}
Our second set of experiments is conducted to explore whether our method can outperform existing one-shot data-free pruning methods\footnote{Models are pruned by up to 50\% as it is the upper limit of one-shot pairwise pruning approaches by default.}.
We compare its performance with that of the saliency-based one-shot pruning, which is a commonly used approach to perform data-free neural network pruning~\cite{srinivas2015data,liu2019rethinking}.
We note that our evaluation focuses on the comparison of data-free pruning techniques, and we refer the reader to the existing study~\cite{han2015learning} that compares performance between data-driven and data-free pruning techniques.

We reuse the same four models and take the saliency-based one-shot pruning as the baseline.
The improvement of accuracy is equal to the growth of accuracy of the pruned model produced by our method relative to the one pruned by the baseline method.
The improvement of robustness is calculated as the growth in the number of robust instances observed from running our method relative to the baseline. We present our benchmarking outcomes in Fig.~\ref{fig:improvement}.

We find both our method and the baseline can perform almost a lossless pruning for model \#1. That is because the task (i.e., binary classification for credit card fraud detection) is comparably simple and the model architecture is designed with too much redundancy. Besides that, our method outperforms the saliency-based one-shot pruning in all remaining three settings.
In the experiments on MNIST and CIFAR-10 models (i.e., models \#3-4), our method achieves a significant improvement in both robustness preservation (up to 42\% in model \#3) and accuracy (up to 66\% in model \#4).
In model \#2, our method achieves much more significant improvement with up to 3x (observed after 46\% of hidden units have been pruned), indicating that by applying our method, the model can be further compressed to preserve a similar performance.

We also observe the change in accuracy and robustness of a model is dependent on the utilization of its hidden units.
In model \#2, the improvement of accuracy starts declining after 45\% units are pruned. By referring to Fig.~\ref{fig:decay}, we find that the model fidelity also starts declining at almost the same time. This shows that a 45\% pruning eliminates most computationally negligible parameters in model \#2 and any further model decompression would come at the expense of sacrifice of model fidelity.
A similar phenomenon can also be observed in the robustness improvement in model \#4. The improvement of our pruning starts declining after 45\% units are pruned, but this phenomenon does not appear in model \#3.
Although both models \#3 and \#4 are designed for similar tasks, i.e., MNIST and CIFAR-10, model \#3 is trained as a fully connected MLP, while model \#4 is trained as a CNN with only a small portion of its units are fully connected.
This makes the former model contains more computationally negligible parameters than the model of CIFAR-10, and therefore the pruning of the former yields less impact on the robustness than the latter one.

%% file: sections/table_of_dataset_and_models_single_col.tex

\begin{table}[t]
	\footnotesize 
	\begin{center}
	\def\arraystretch{1.1}
	\setlength{\tabcolsep}{2.5pt}
	\caption{\label{tab:table-datasets} Datasets and models used in evaluation}
	\begin{tabular}{cllr}
		\hline
		\multicolumn{2}{l}{\textbf{Models \& Datasets}}     & \multicolumn{1}{l}{\textbf{Model Architecture}} & \multirow[t]{2}{*}{\textbf{\makecell[lt]{Num. of Parameters}}} \\ \hline
		1 & Credit Card~\textsuperscript{$\dagger$} & 4-layer MLP & 6,145 \\ \hline
		2 & KDD Cup'99~\textsuperscript{$\ddagger$} & 6-layer MLP & 245,655  \\ \hline
		3 & MNIST & 5-layer MLP & 125,898 \\ \hline
		4 & CIFAR-10 & 13-layer CNN (3 FC layers) & 753,866 \\ \hline
	\end{tabular}
\end{center}
\vspace{-8pt}
\begin{flushleft}
	\begin{footnotesize}
		\textsuperscript{$\dagger$} Available at \url{https://www.kaggle.com/mlg-ulb/creditcardfraud}. \\
		\textsuperscript{$\ddagger$} Available at \url{http://kdd.ics.uci.edu/databases/kddcup99/kddcup99.html}. \\
	\end{footnotesize}
\end{flushleft}
\end{table}

%% file: sections/table_of_elapsed_time.tex
\begin{table}[t]
	\footnotesize  
	\begin{center}
	\def\arraystretch{1.1}
	\setlength{\tabcolsep}{4pt}
	\centering
	\captionsetup{justification=centering,margin=0cm}
	\caption{\label{tab:table-elapsed-time} Time consumed in pruning 80\% of FC parameters in the most conservative setting}
	\vspace{-5pt}
	\begin{tabular}{lrr}
		\hline
		\textbf{Model} &
			\textbf{\makecell[l]{Batch size (per layer)}} &
			\textbf{\makecell[l]{Elapsed time (secs)}} \\\hline
		\#1 (Credit Card) & 3.13\%       & 30.5 \\ \hline
		\#2 (KDD Cup'99)        & 3.13\%       & 76.7 \\ \hline
		\#3 (MNIST)       & 1.56\%       & 435.4 \\ \hline
		\#4 (CIFAR-10)    & 1.56\%       & 265.3 \\ \hline
	\end{tabular}
\end{center}
\vspace{-15px}
\end{table}

%% file: sections/limitations.tex
Our work carries several limitations that should be addressed in the future.

First, our method is primarily designed for fully connected components of a neural network model. 
Fully connected layers are fundamental components of deep learning and have been increasingly used in state-of-the-art designs such as MLP-Mixer~\cite{tolstikhin2021mlpmixer}.
Nevertheless, our method may be limited when applied to models with convolutional and relevant layers (e.g., pooling and normalization) playing a major role.
We still need to explore more regarding how to effectively prune diverse models that are not built in conventional fully connected architecture, such as transformer models.

Second, our pruning heavily relies on interval arithmetic in approximating the valuation of hidden units and pruning impact, so the precision of those intervals determines both the effectiveness and correctness of our method.
When it is applied to a ReLU-only model with a large number of hidden layers,
there may be a magnitude explosion issue during our evaluation of propagated impact on the output layer.
Besides, pruning a model mixed with both convergent (e.g., sigmoid) and non-convergent (e.g., ReLU) activation may be challenging for our method,
because a convergent activation may reduce the quantitative difference from the previous assessment and output a similar result close to $\left(-1,1\right)$ --- this may reduce the effectiveness of our sampling criterion.

We share our insight for future work to mitigate these limitations in two aspects. 
First, the data-free pruning could be extended to more layer types especially those over-parameterized layer types like 2D convolutional layer. Additional pruning criteria may address the first limitation. 
Second, a more precise interval approximation or refinement technique could be applied to optimize the pruning criteria. By doing this the magnitude explosion issue of the propagated impact on the output layer may be relieved.

%% file: sections/review.tex

\paragraph{Unstructured Pruning \& Structured Pruning}
Existing pruning approaches can be classified into two classes, namely \textit{unstructured pruning} and \textit{structured pruning}~\cite{liu2019rethinking}.
Unstructured pruning is also known as \textit{individual weight pruning}, which is performed to cut one specific (redundant) parameter at a time. 
It typically prunes weight parameters based on a Hessian of the loss function.
Existing studies that can be categorized as unstructured pruning include~\cite{han2015learning,molchanov2017variational}.

Structured pruning is proposed to prune a model at the hidden unit, channel, or even layer level.
Hu \textit{et al.}~\cite{hu2016network} proposed a channel pruning technique according to the average percentage of zero outputs of each channel, while another study by Li \textit{et al.}~\cite{li2016pruning} presented a similar channel pruning but according to the filter weight norm.
Besides that, there is another common approach discussed in~\cite{yu2018nisp,molchanov2017pruning} that prunes a component with the least influence to the final loss.
He \textit{et al.}~\cite{he2017channel} and Luo \textit{et al.}~\cite{luo2017thinet} proposed channel pruning based on consequential feature reconstruction error at the next layer.
Srinivas and Babu~\cite{srinivas2015data} introduced a data-free pruning method based on saliency, which performs pruning independently of the training process and as the result, does not need to access training data.
Recent work also includes~\cite{suau2020filter} that considers the inter-correlation between channels in the same layer. Another work by Chin \textit{et al.}~\cite{chin2018layer} proposes a layer-by-layer compensate filter pruning algorithm.

\paragraph{In-training Pruning \& Post-training Pruning}
On the other hand, depending on when the parameters' pruning is performed, we can also categorize existing pruning strategies as either \textit{in-training pruning} or \textit{post-training pruning} (also known as data-free pruning)~\cite{tanaka2020pruning}.
Besides a few papers that discuss post-training pruning~\cite{ashouri2019retraining,srinivas2015data}, most existing studies are implemented as in-training pruning.

One recent in-training pruning approach named \textsc{SNIP}~\cite{lee2018snip} achieves single-shot pruning based on connection sensitivity and has been exhaustively compared with existing techniques.
Hornik \textit{et al.}~\cite{hornik1989multilayer} investigated a data-agnostic in-training pruning that proposes a saliency-guided iterative approach to address the layer-collapse issue.
In-training pruning gives us a chance to fine-tune or even retrain the pruned network with the original dataset, and therefore it is capable to prune a larger portion of the neural network without worrying about a severe impact on the model fidelity (e.g., accuracy and loss).
A recent empirical study performed by Liebenwein \textit{et al}.~\cite{liebenwein2021lost} reveals the robustness could be well preserved during the mainstream in-training pruning.
Even though, post-training such as~\cite{srinivas2015data,ashouri2019retraining} are useful to reduce the size of a pre-trained ready-to-use neural network model from a user's perspective.
The effectiveness of post-training pruning beyond accuracy, such as robustness preservation, has yet to be well studied.

%% file: sections/conclusion.tex
In this work, we propose a supervised pruning method to achieve data-free neural network pruning with robustness preservation.
Our work aims to enrich the application scenarios of neural network pruning, as a supplementary of the state-of-the-art pruning techniques that request data for retraining and fine-tuning.
With our proposed sampling criterion, we take advantage of simulated annealing to address the data-free pruning as a stochastic optimization problem.
Through a series of experiments, we demonstrate that our method is capable of preserving robustness while substantially reducing the size of a neural network model, and most importantly, without a significant compromise in accuracy.
Our evaluation shows that our pruning can generalize on diverse types of models and datasets.
We remark that the model pruning in the context of data-freeness is a practical problem, and more future studies are desirable to cope with the challenges we report in this work.